\documentclass{article}
\usepackage{ijcai24}

\usepackage{times}
\usepackage{soul}
\usepackage{url}
\usepackage[hidelinks]{hyperref}
\usepackage[utf8]{inputenc}
\usepackage[small]{caption}
\usepackage{graphicx}
\usepackage{amsmath}
\usepackage{amsthm}
\usepackage{booktabs}
\usepackage{algorithm}
\usepackage{algorithmic}
\usepackage[switch]{lineno}
\usepackage{multirow}
\usepackage{graphicx}
\usepackage{array}
\usepackage{adjustbox}
\usepackage{xcolor}
\usepackage{setspace}

\title{Transferring Core Knowledge via Learngenes}

\author{
Fu Feng$^{1,2}$
\and
Jing Wang$^{1,2*}$  \and
Xin Geng$^{1,2}$\footnote{Co-corresponding author}
\affiliations
$^1$School of Computer Science and Engineering, Southeast University, Nanjing, China\\
$^2$Key Laboratory of New Generation Artificial Intelligence Technology and Its Interdisciplinary Applications (Southeast University), Ministry of Education, China\\
\emails
\{fufeng, wangjing91, xgeng\}@seu.edu.cn}

\begin{document}

\maketitle

\begin{abstract}
    The pre-training paradigm fine-tunes the models trained on large-scale datasets to downstream tasks with enhanced performance. It transfers all knowledge to downstream tasks without discriminating which part is necessary or unnecessary, which may lead to negative transfer. In comparison, knowledge transfer in nature is much more efficient. When passing genetic information to descendants, ancestors encode only the essential knowledge into genes, which act as the medium. Inspired by that, we adopt a recent concept called ``learngene'' and refine its structures by mimicking the structures of natural genes. We propose the Genetic Transfer Learning (GTL)---a framework to copy the evolutionary process of organisms into neural networks. GTL trains a population of networks, selects superior learngenes by tournaments, performs learngene mutations, and passes the learngenes to next generations. Finally, we successfully extract the learngenes of VGG11 and ResNet12. We show that the learngenes bring the descendant networks instincts and strong learning ability: with 20\% parameters, the learngenes bring 12\% and 16\% improvements of accuracy on CIFAR-FS and miniImageNet. 
    Besides, the learngenes have the scalability and adaptability on the downstream structure of networks and datasets. Overall, we offer a novel insight that transferring core knowledge via learngenes may be sufficient and efficient for neural networks.
\end{abstract}

\section{Introduction}
The escalating number of parameters in neural networks has led to an exponential growth in requisite training data~\cite{yu2020self}. Consequently, the pre-training paradigm fine-tunes the models trained on large-scale datasets to the specific tasks with small-scale datasets~\cite{zoph2020rethinking,chakraborty2022efficient}. It leverages the learned knowledge from the large-scale datasets to these specific tasks, which can accelerate the training speed and enhance the model performance~\cite{chen2020adversarial,li2021improve}. 
Moreover, there is a prevailing effort to leverage and preserve all acquired knowledge in neural networks. 
Techniques like knowledge distillation (KD) aim to maximize the transfer of knowledge from teacher to student models~\cite{wang2021knowledge,huang2022knowledge}, while model compression endeavors to retain as much knowledge as possible during size reduction~\cite{frantar2023sparsegpt,yu2020self}. 

\begin{figure}[tb]
  \centering
  \includegraphics[width=\linewidth]{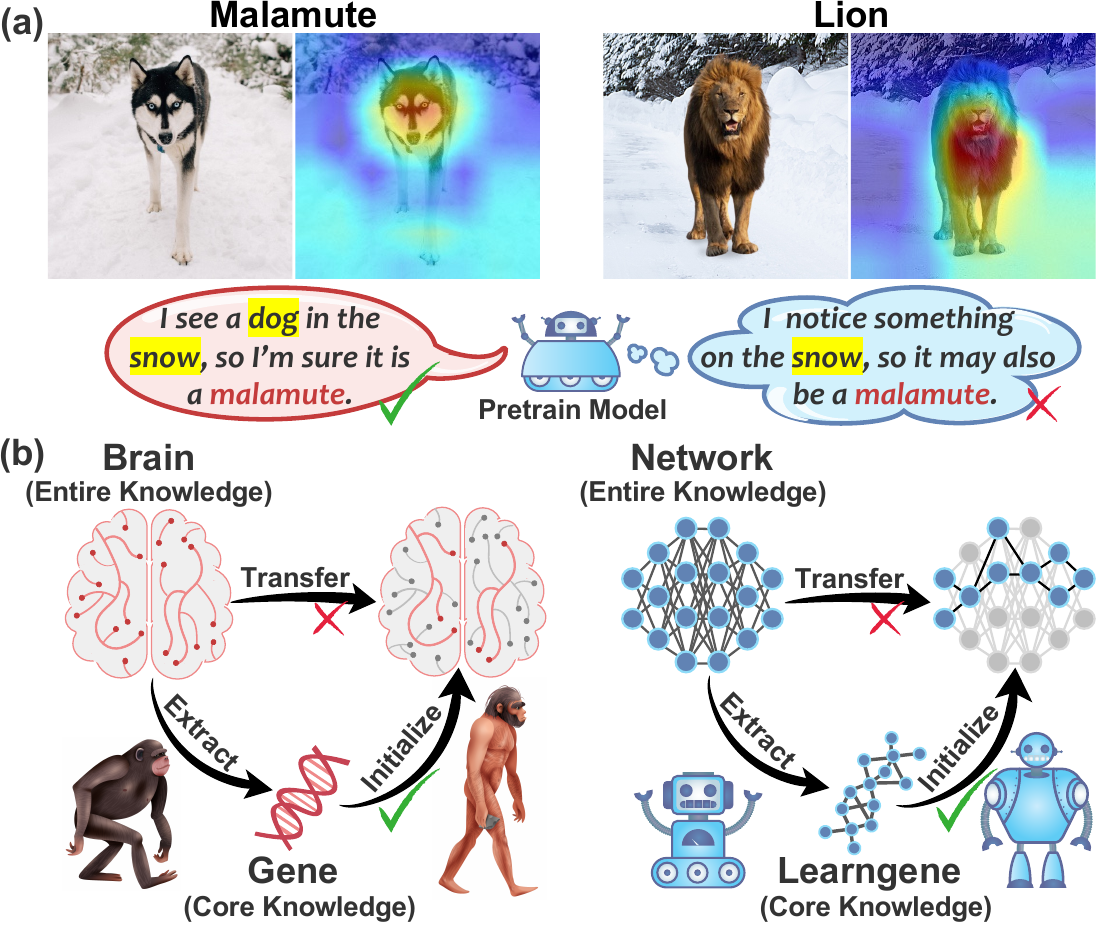}
  \caption{(a) Transferring the entire knowledge may be redundant or negative. (b) Leveraging the learngenes to transfer the core knowledge to descendant networks inspired by the genes in nature.}
  \label{fig:motivation}
\end{figure}

These works transfer all knowledge to downstream tasks without discriminating which part is necessary or unnecessary. Indeed, knowledge transfer is not necessarily better with more, as excessive knowledge transfer may result in redundancy and negative transfer~\cite{rosenstein2005transfer,wang2019characterizing}. 
As depicted in Figure \ref{fig:motivation}(a), once a pre-trained model has mastered the knowledge about snow when pre-training, it tends to strengthen the association between malamutes and snow during fine-tuning (sled dogs are often active in snow), and therefore misclassify a lion in the snow as a malamute due to the redundancy of knowledge about snow.
In contrast, nature takes a different approach to knowledge transfer. When passing the genetic information to descendants, the ancestors do not transmit the whole lifetime's knowledge in their brains and, instead, encode only the essential knowledge vital for survival into genes, which act as the medium for knowledge transfer~\cite{bohacek2015molecular,waddington1942canalization}. 
The descendants inheriting the genes have instincts, which enable them with strong learning ability to acquire new knowledge in their environments~\cite{wong2015behavioral,sih2011evolution}.

We may raise the question that \textit{can neural networks benefit from the way (via genes) of knowledge transfer in nature}? To this end, we adopt the concept of ``learngene'' proposed by \cite{feng2023genes,wang2023learngene}---the inheritable ``genes'' of neural networks regarding learning ability, which consists of a core subset of neural networks (e.g., layers). In this paper, we refine the learngenes by mimicking the evolution process of natural genes.  
As illustrated in Figure \ref{fig:motivation}(b), the ancestor networks (i.e., well-trained networks for knowledge provision) transfer knowledge via the inheritable learngenes to descendant networks (i.e., random initialized networks for knowledge inheritance). Like natural genes, the learngenes condense only the core knowledge, emphasizing fundamental local features that concentrate on the classification object itself, while disregarding irrelevant or redundant information. Thus, the learngenes enable descendant networks to fast adapt to diverse environments, and are much more flexible and efficient compared to the transfer of entire networks.

To extract the learngenes from neural networks, we model the learngenes as neural connections (i.e., continuous feature mappings in the unit of channels within kernels) and propose Genetic Transfer Learning (GTL) that is a framework adapted and refined from Genetic Reinforcement Learning (GRL) \cite{feng2023genes}, specifically designed for supervised learning tasks. GTL copies the evolutionary process of organisms into neural networks. First, GTL partitions the datasets into small classification tasks to simulate the survival environments of neural networks. 
Second, to simulate the natural selection and inheritance, GTL trains a population of neural networks and applies tournaments to select superior learngenes, which can be passed to next generations. Third, GTL performs gene mutations, allowing the learngenes to adaptively adjust their structures during the evolutionary process for more effective storage of acquired core knowledge.

After 250 generations of evolution, we successfully extracted the learngenes from the neural networks of VGG and ResNet. Despite comprising only approximately 20\% of the total network parameters, the learngenes bring 12\% and 16\% improvements of accuracy for the validation and novelty classes of CIFAR-FS and $\textit{mini}$ImageNet, respectively, compared with learning from scratch. 
Furthermore, the learngenes bring the neural networks the instincts and strong learning ability, requiring minimal data and parameter updates to enable the condensed core knowledge within the learngenes to adapt to the features of the current datasets.
They also exhibit notable scalability and adaptability to diverse data types and network structures in downstream tasks.

Our main contributions are as follows: 
1) We refine the structure of the learngenes and use the learngenes to condense and transfer core knowledge in neural networks. Compared to the pre-training paradigm, our work provides an alternative method for knowledge transfer. 2) We propose GTL, a framework for large-scale neural network evolution in supervised learning. Leveraging GTL, we present the process of evolution, mutation, and inheritance of the learngenes, and successfully extract the learngenes from convolutional neural networks (CNNs).
3) We validate the advantages of the learngenes. The learngenes bring instincts and strong learning ability to descendant networks with the flexibility of parameters. The learngenes also exhibit scalability and adaptability across diverse network structures and downstream tasks.

\section{Related Work}
\paragraph{Transfer Learning.} 
Transfer learning aims to effectively convey knowledge from neural networks trained in the source domain to the target domain, thus facilitating knowledge acquisition of target neural networks~\cite{zhuang2020comprehensive,iman2023review}.
In traditional transfer learning, knowledge is transferred through a pre-trained model with an identical structure to the target network~\cite{he2019rethinking,zoph2020rethinking}. This approach involves transferring entire knowledge from the source domain, potentially introducing redundant information that impacts the learning of neural networks in the target domain.
While knowledge distillation relaxes structural constraints during knowledge transfer, its fundamental goal remains the comprehensive transmission of knowledge from teacher models to student models~\cite{hinton2015distilling}.
In contrast, the learngenes adopt a distinct knowledge transfer strategy inspired by the knowledge transfer mode of genes in nature. 
The learngenes selectively transfer core knowledge within neural networks rather than entire knowledge, represented as specific neural circuits comprising neuron connections. 
The structural flexibility of learngenes enables them to adapt to the target model's architecture while efficiently transferring core knowledge.

\paragraph{Evolutionary Learning.}
Evolutionary Learning draws inspiration from natural evolution to address optimization problems in a stochastic manner~\cite{telikani2021evolutionary}. Thus, algorithms in Evolutionary Learning inevitably introduce the concepts akin to ``genes'', exemplified by the ``genomes'' or ``chromosomes'' employed in genetic algorithms~\cite{zhou2019evolutionary}. It is essential to note that, in genetic algorithms or other related Evolutionary Learning algorithms~\cite{stanley2002evolving,stanley2009hypercube,mirjalili2019genetic,sivanandam2008genetic,mirjalili2020genetic}, ``genome'' or ``chromosome'' serve as representations for candidate solutions in optimization. In contrast, the ``learngenes'' in this paper function as mediums for transferring core knowledge.
Evolutionary Learning algorithms have found successful applications in neural networks, involving the learning of weight parameters, hyperparameters, or architectures, where the evolution is still designed for optimization~\cite{unal2022evolutionary,mishra2023survey,darwish2020survey,zhou2021survey}. 
While the evolution in this paper primarily serves as a mechanism for the inheritance and evolution of learngenes. The neural networks are employed solely to simulate the organisms, with their parameters optimized through gradient descent for learning. 
In essence, aside from potential literal confusion due to similar nouns, the exploration of learngenes and evolution in this article bears no other connection to Evolutionary Learning.

\section{Methods}
\subsection{Form of the Learngenes}
In the biological neural networks of our brain, several innate neural circuits are established at birth under the guidance of genes. So newborns have the instincts with strong learning ability, which learn fast in their own environments~\cite{wei2021neural,luo2021architectures,zador2019critique}. 
Since artificial neural networks simulate biological neural networks from the perspective of information processing, we abstract the learngenes into discrete neural circuits within artificial neural networks. Specifically, in convolutional neural networks (CNN), learngenes are represented as channels within convolutional kernels, preserving the continuity mapping of features.

For a CNN with $n_L$ layers, it can be symbolized in terms of channels as $\mathcal{N} = \{\mathcal{C}_{c,k,l} | c\in [1, n^l_C], k\in [1, n^l_K], l\in [1, n_L] \}$, where $l$, $k$ and $c$ are indices of the layer, kernel and channel, respectively. Here, $\mathcal{C}_{c,k,l}$ represents the $c$-th channel of the $k$-th kernel in the $l$-th layer. $n^l_K$ denotes the number of kernels, and $n^l_C$ represents the number of channels for each kernel in the $l$-th layer. 

\begin{figure}[tb]
  \centering
  \includegraphics[width=\linewidth]{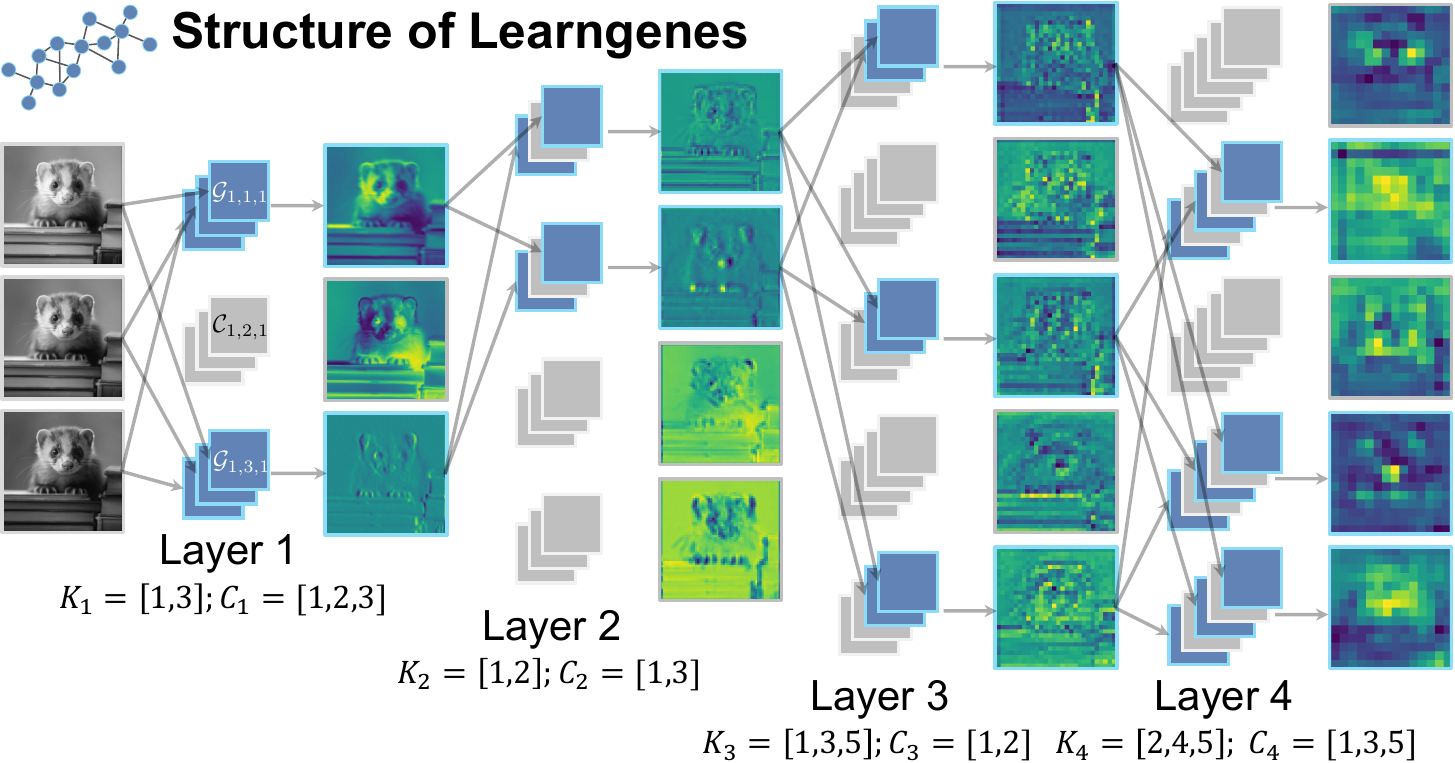}
  \caption{Structure of the learngenes, which are several complete neural circuits in the unit of channels (colored blue) within kernels.}
  \label{fig:form}
\end{figure}

Since the learngenes in CNNs are comprised of channels within kernels, we formulate the learngene as $\mathcal{G} = \{ \mathcal{G}_{c,k,l} | k \in K_l, c \in C_l, l\in [1, n_L] \}$, where $\mathcal{G}_{c,k,l}$ means that the $\mathcal{C}_{c,k,l}$ is a part of the learngene $\mathcal{G}$. 
$K_l$ and $C_l$ are sets of indices of the kernels and channels associated with the learngene in $l$-th layer, respectively (Figure \ref{fig:form}). 
In a CNN, the correlation between the number of kernels and channels in consecutive layers is explicit (i.e., $n^l_K=n^{l+1}_C$). Thus, to preserve the continuous mapping of features within the learngenes, we maintained structural consistency by setting $K_l = C_{l+1}$.

\begin{figure*}[tb]
  \centering
  \includegraphics[width=0.918\linewidth]{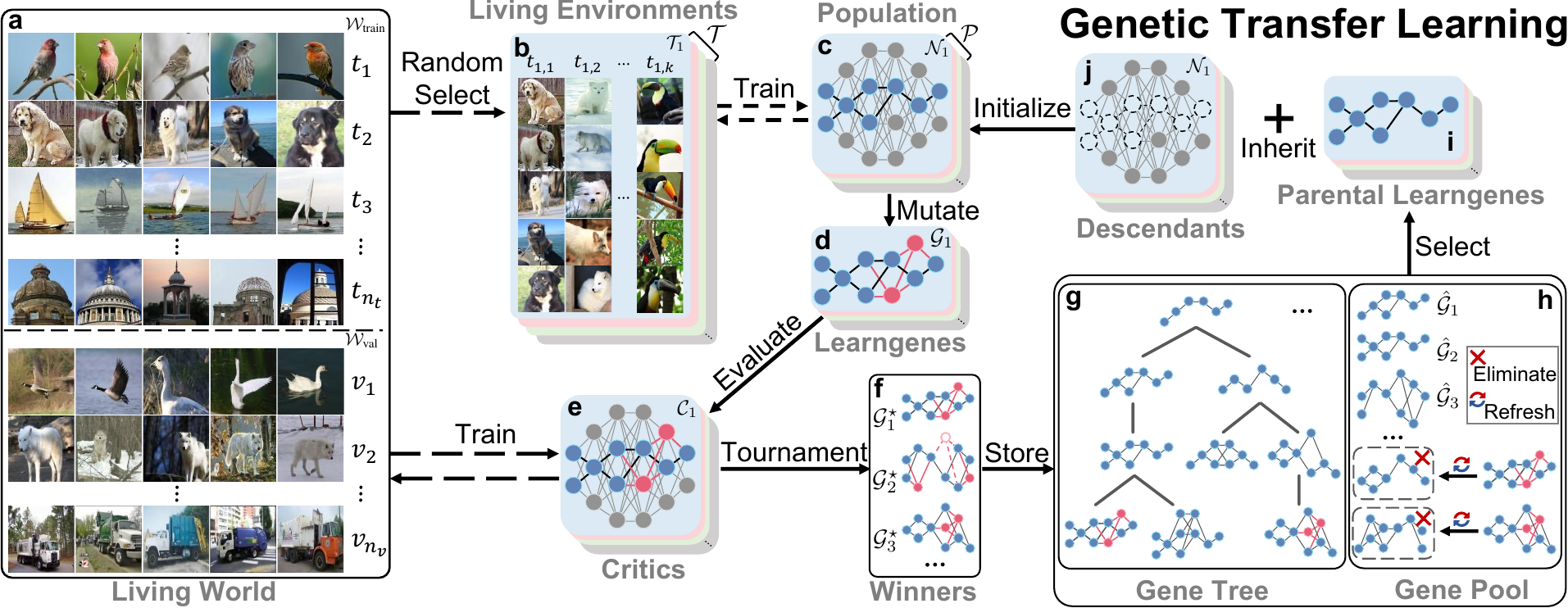}
  \caption{The framework of Genetic Transfer Learning (GTL), which are used to condense core knowledge and extract the learngenes.}
  \label{fig:GTL}
\end{figure*}

\subsection{Mutation of the Learngenes}
Gene mutations promote biological evolution, describing evolution as the accumulation of small dominant mutations.
To align the evolution of the learngenes with the natural genetic processes, we define mutations within the learngenes.

The mutations of the learngenes fundamentally entail structural modifications, such as the augmentation or reduction of kernels and channels within a specific layer. In this study, due to the alignment of kernels and channels between adjacent layers (i.e., $K_l = C_{l+1}$), we focus on mutations on kernels, subsequently adjusting the corresponding channels. For a single mutation in the learngenes, we consider its possibility independently across all layers, with each layer exhibiting a probability $p_m$ for undergoing this mutation. The likelihood of either increasing or decreasing a specific kernel in each layer is then computed as follows:
\begin{equation}
    p^{\scriptscriptstyle +}_l = \alpha \cdot \frac{|K_l|}{n_K^l - |K_l|} \quad \text{and}\quad
    p^{\scriptscriptstyle -}_l = 1 - p^{\scriptscriptstyle +}_l
\end{equation}
where $p^{\scriptscriptstyle +}_l$ and $p^{\scriptscriptstyle -}_l$ represent the probabilities of increasing and decreasing a kernel in $l$-th layer of the learngene. 

A single mutation of the learngenes constitutes a continuous process, which may occur across multiple layers. Each layer has the potential for multiple kernel changes, as outlined in Algorithm \ref{alg:algorithm}.

\begin{algorithm}[tb]
    \caption{Mutation of the Learngene}
    \small
    \label{alg:algorithm}
    \textbf{Input}: Learngene $\mathcal{G}$
    \begin{algorithmic}[1]
        \FOR{$l = 1$ to $n_L$}
            \STATE Randomly generate $r \sim U(0, 1)$.
            \WHILE{$r \leq p_m$}
                \STATE Randomly generate $s \sim U(0, 1)$. 
                \IF{$s \leq p^{\scriptscriptstyle +}_l$}
                    \STATE Randomly select $k$ from $[1, n^l_K] - K_l$.
                    \STATE $K_l \leftarrow K_l \cup \{k\}$ and $C_{l+1} \leftarrow C_{l+1} \cup \{k\} $
                \ELSE
                    \STATE Randomly select $k$ from $K_l$.
                    \STATE $K_l \leftarrow K_l - \{k\}$ and $C_{l+1} \leftarrow C_{l+1} - \{k\} $
                \ENDIF
            \STATE Randomly generate $r \sim U(0, 1)$.
            \ENDWHILE
        \ENDFOR
    \end{algorithmic}
\end{algorithm}

\subsection{Inheritance of the Learngenes}
The natural divergence of genes ensures a degree of scalability, for instance, allowing cats and lions to share ancestral feline genes.
Similarly, learngenes exhibit adaptability to the structures of descendant networks, accommodating variations in depth, width and architecture (see \textbf{Appendix B}). 
In the evolution process of extracting the learngenes, we maintained a consistent population structure for simplification. 

\subsubsection{Completing Missing Channels within Kernels}
The kernels in the learngenes are incomplete and contain only a subset of channels, which inevitably introduces randomly initialized channels when inheriting the learngenes. This may disrupt the core features extracted by learngene kernels. To mitigate this problem, before inheriting the learngenes, we will fill with $\mathbf{0}$ to the missing channels in each kernel of the learngenes. This ensures that in convolution operations, these missing channels do not affect the core features already extracted by the learngenes and do not compromise the kernels' capacity to learn new features.

\subsubsection{Adjusting the Kernel and Channel Positions}
When a descendant network shares the same architecture as the ancestry networks, it just needs to simply replace its randomly initialized kernels with the learngenes. 
However, when the descendant network is narrower, certain indices of learngene kernels and channels may exceed the descendant network's maximum index.
Given that learngenes maintain the adaptability of CNNs, whose kernels and channels can be sorted to adapt to neural networks of different widths under the condition of $K_l = C_{l+1}$.

\subsubsection{Expanding the Depth of the Learngenes}
In addition to accommodating varying widths in descendant networks, learngenes can also initialize a network with greater depth. For a network with $n_L^d$ layers while the learngene with $n_L^a$ layers ($n_L^a\!<\!n_L^d$), we need to add $n_L^d \!-\! n_L^a$ \textit{partial identity mapping} layers $l_{\text{pim}}$ to the learngenes. 

For a $l_{\text{pim}}$ between the $l$-th and $(l\text{+1})$-th layer, it extends the original feature mapping path of the learngenes, transforming the feature mapping path from $l \!\rightarrow\! l\text{+1}$ to $l \!\rightarrow\! l_{\text{pim}} \!\rightarrow\! l\text{+1}$. Not only does $l_{\text{pim}}$ align the number of layers in the learngene with that of the descendant network, but it also seamlessly transfers core features extracted from the $l$-th layer to $(l\text{+1})$-th layer, ensuring the continuous mapping of core features within the learngene. For the structure of $l_{\text{pim}}$ between $l$ and $l\text{+1}$, the number of its kernel and corresponding channels are $n^{l_{\text{pim}}}_K\!=\!n^{l_{\text{pim}}}_C\!=\!n^l_K$. For the $k$-th kernel in $l_{\text{pim}}$, if $k\!\in\! K_l$, we initialize it as follows, and others are random initialized.
\begin{equation}
    \mathcal{C}_{c,k,\text{pim}} = \left\{
            \begin{array}{ll}
              \mathbf{\mathring{1}} & \text{if } c = k \\
              \mathbf{0} & \text{otherwise }
            \end{array};
         \right.
    \mathbf{\mathring{1}}_{3 \times 3} = [\begin{smallmatrix} 0 & 0 & 0 \\ 0 & 1 & 0 \\ 0 & 0 & 0 \end{smallmatrix}]
    \label{equ:pim}
\end{equation}

\subsection{Extraction of the Learngenes}
The genes in nature have undergone 3.5 billion years of evolution, culminating in the biological intelligence observed today~\cite{braga2017emperor,oro2004evolution}. Inspired by that, we extract the learngenes by simulating the large-scale evolution of the organisms in neural networks. Adapting the GRL framework \cite{feng2023genes}, designed for agents evolution in reinforcement learning, we tailored and extended it as Genetic Transfer Learning (GTL), to concurrently train networks on image classification tasks while evolving the learngenes across multiple generations, as shown in Figure \ref{fig:GTL}.

Each generation starts with a population of $n_p$ neural networks, wherein each network inherits the learngenes from previous generations and is randomly assigned a task. Evolution starts after the training of all $n_p$ neural networks, with $s$ networks randomly selected to participate in a tournament. In each tournament, the winner has a chance to enter the Gene Pool for subsequent generations. After completing of all tournaments, a new generation runs in a nested cycle of learning and evolution. At the end of evolution, the learngenes in Gene Pool are our final extracted learngenes. 

\subsubsection{Training the Population of Neural Networks}
The living world of the neural networks, denoted as $\mathcal{W} = \mathcal{W}_{\text{train}} +  \mathcal{W}_{\text{val}}$, comprises of a total $n = n_t + n_v$ classes. Here $\mathcal{W}_{\text{train}} = \{t_1, t_2, ..., t_{n_t}\}$ serves as the environments for the population's survival in generational evolution, and $\mathcal{W}_{\text{val}} = \{v_1, v_2, ...,  v_{n_v} \}$ serves as unseen environments for learngene performance evaluation (Figure \ref{fig:GTL}a).

The neural network population $\mathcal{P}=\{\mathcal{N}_1, \mathcal{N}_2, ..., \mathcal{N}_{n_p}\}$ is generated in each generation (Figure \ref{fig:GTL}c). Each network $\mathcal{N}_{i}$ randomly select $k$ classes from $n_t$ classes in $\mathcal{W}_{\text{train}}$, constructing a $k$-classification task $\mathcal{T}_i$ as its survival environment (Figure \ref{fig:GTL}b). As evolution progresses, survival environments become more complex, which can be achieved by increasing the value of $k$ to construct more challenging classification tasks.

\subsubsection{Selecting the Superior Learngenes}
After the training of neural networks in one generation, we extract the learngene $\mathcal{G}_i$ from $\mathcal{N}_i$ and mutate $\mathcal{G}_i$ based on Algorithm \ref{alg:algorithm} (Figure \ref{fig:GTL}d).
Then, each learngene $\mathcal{G}_i$ initializes a critic network $\mathcal{C}_i$, and $\mathcal{C}_i$ will be trained in $\mathcal{W}_{\text{val}}$ to evaluate the performance of $\mathcal{G}_i$, whose accuracy will be used as the score $s_i$ of the $\mathcal{G}_i$  (Figure \ref{fig:GTL}e). 

Next, superior learngenes are selected based on their scores with the opportunity to produce descendants. To preserve diversity during evolution, we employ tournaments to select superior learngenes.
Each tournament randomly selects $\delta$ learngenes from $\mathcal{P}$ (without replacement), and the learngene with the highest score is the winner. $\mathcal{G}^{\star} = \{\mathcal{G}_1^{\star}, \mathcal{G}_2^{\star}, ..., \mathcal{G}_{n_w}^{\star} \}$ ($n_w=7$ with $n_p=20$ and $\delta=3$) represents superior learngenes selected by tournaments in a generation (Figure \ref{fig:GTL}f).

\subsubsection{Storing the Learngenes and Their Kinship}
Following the tournament selection of $\mathcal{G}^{\star}$, the Gene Pool (GP) is utilized to store these superior learngenes as candidate parents for generating descendants (Figure \ref{fig:GTL}h). Besides, the Gene Tree (GT) is employed to record the kinship of these learngenes throughout the entire evolution process, archiving the ancestry learngenes (Figure \ref{fig:GTL}g).

In the initial generation, the learngenes are formed by randomly selecting $c \cdot n_l$ kernels in each layer from the ancestors. $\text{GP} = \{\hat{\mathcal{G}}_1, \hat{\mathcal{G}}_2, ..., \hat{\mathcal{G}}_{\rho _{\text{max}}} \}$ can store up to $\rho_{\text{max}}$ learngenes, which was initialized by $\mathcal{G}^{\star}$ of the initial generation. In subsequent generations, only $\varepsilon$ learngenes in $\mathcal{G}^{\star}$ have the chance to be  to be added to the GP, preventing significant changes.
Nodes in the GT are the learngenes (currently or previously) in the GP, with root nodes being the learngenes in the initial GP. Each generation adds the learngenes stored in GP as a new leaf node in GT, where the path length between nodes reflects the closeness of kinship between the learngenes. 

\subsubsection{Updating the Scores of the Learngenes}
We update the scores of the learngenes in the GP after selecting $\mathcal{G}^{\star}$ to preserve the excellence and continuity of ancestors. For $\mathcal{G}^{\star}_i$, we start from a leaf node of the GT (i.e., the parent learngene of $\mathcal{G}^{\star}_i$) and backtrack to the root node. If the ancestry node (i.e., ancestry learngene) in the path is in the current GP, the score of this learngene $s_{\text{anc}}$ will be updated: 
\begin{equation}
    \hat{s}_{\text{anc}} \leftarrow \hat{s}_{\text{anc}} + \eta ^{\tau} s_i
\label{equ:update}
\end{equation}
where $s_{\text{anc}}$ is the score of the ancestry learngene, $\eta$ is the parental decay coefficient, and $\tau$ is the path length between the ancestry node and the leaf node on the GT.

Upon completing the update of learngene scores and the replacement of the learngenes in the GP, the learngenes of the next generation of neural networks $\mathcal{P}$ is generated based on the probability calculated by
\begin{equation}
    \hat{p}_i = \frac{\hat{s}_i}{\sum_{i=1}^{\rho_{\text{max}}} \hat{s}_i}
\label{equ:prob}
\end{equation}
where $\hat{p}_i$ is the probability of $\hat{\mathcal{G}}_i$ being selected as the parent learngene with score $\hat{s}_i$ (Figure \ref{fig:GTL}i). Then, a new round of evolution starts (Figure \ref{fig:GTL}j).

\section{Experiments}
\subsection{Experimental Setting}

\textbf{Datasets.} 
CIFAR-FS~\cite{bertinetto2018metalearning} and $\textit{mini}$ImageNet~\cite{vinyals2016matching} are subsets randomly sampled from CIFAR-100 and ImageNet by the same criteria, respectively. Each dataset comprises 100 object classes with a total of 60,000 images, categorized into training, validation, and novelty classes, with class numbers of 64, 16, and 20. In our experiments, the training, validation, and novelty classes are used to simulate the living environments of the neural networks (i.e., $\mathcal{W}_{\text{train}}$), evaluate the performance of the learngenes during evolution (i.e., $\mathcal{W}_{\text{val}}$), and test the advantages of the extracted learngenes, respectively.\\ 
\textbf{Network Architectures.}
The images in CIFAR-FS and $\textit{mini}$ImageNet have dimensions of $\text{32} \times \text{32}$ and $\text{84} \times \text{84}$, respectively. To accommodate the dataset complexities and diverse ancestral network structures, we adopt VGG11 and ResNet12 as the neural network architectures evolving on CIFAR-FS and $\textit{mini}$ImageNet, respectively.

\subsection{Learngenes Extracted in Evolution Process}
\label{sec:extract learngene}
\begin{figure}[tb]
  \centering
  \includegraphics[width=\linewidth]{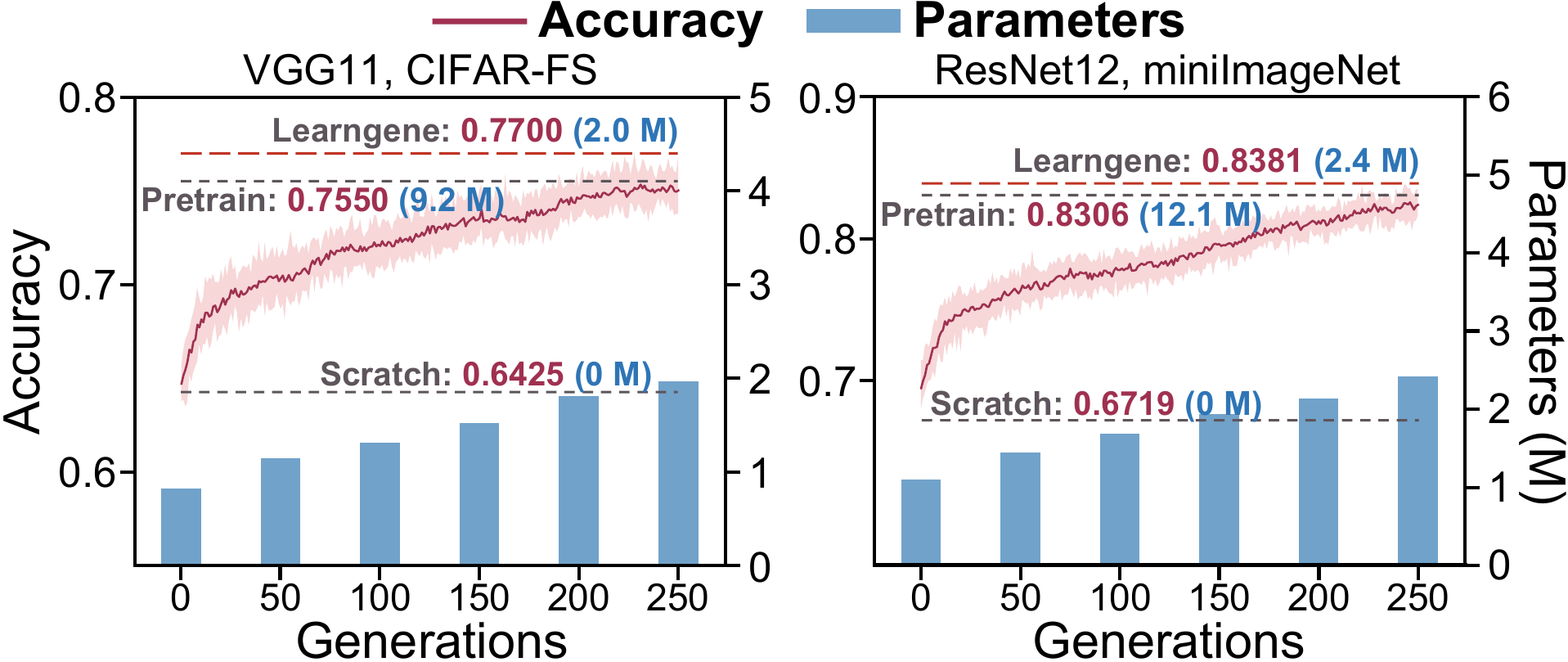}
  \caption{The parameter quantity of the learngenes (i.e., blue bars) and the average (with max and min) accuracy on validation classes of networks in population (i.e., red curves) during evolution. Black lines are accuracy (with the number of transferred parameters) of the models trained from scratch and pre-trained on training classes.}
  \label{fig:evolution}
\end{figure}

In biological evolution, the continuous accumulation of small dominant mutations, coupled with natural selection, drives the continuous evolution of genes~\cite{jablonka1998lamarckian,kimura1983neutral}.
In the experiments, we conducted 250 generations of evolution using VGG and ResNet architectures on CIFAR-FS and $\textit{mini}$ImageNet, respectively, as illustrated in Figure \ref{fig:evolution}. The learngenes also exhibited a continuous accumulation of dominant mutations throughout the evolution, which are reflected in the increasing number of learngene parameters over generations. Figure \ref{fig:evolution} and Table \ref{tab:evolution} demonstrate that neural networks inheriting the learngenes progressively achieve higher accuracy in validation classes (i.e., $\mathcal{W}_{\text{val}}$) and novelty classes, as the core knowledge condensed in the learngenes constantly increase during the evolution.

\begin{table}
    \centering
    \resizebox{\linewidth}{!}{
        \begin{tabular}{ccc|cc}
            \hline
             & \multicolumn{2}{c|}{VGG11, CIFAR FS} & \multicolumn{2}{c}{ResNet12, $\textit{mini}$Imagenet}\\
            Learngene & validation & novelty & validation & novelty\\
            \hline
             --- & 64.25 & 67.10 & 67.19 & 65.65\\
             10th & 68.56 & 69.80 & 74.00 & 74.50 \\
             50th & 71.31 &73.45& 76.81 & 76.80\\
             100th & 72.94 & 75.30 & 79.19 & 78.55\\
             160th & 74.25 & 76.45 & 81.37 & 80.70\\
             250th & \textbf{76.81} & \textbf{78.10} & \textbf{83.75} & \textbf{82.45}\\
            \hline
        \end{tabular}
    }
    \caption{Multi-class classification accuracy of the networks on validation and novelty classes of CIFSAR-FS and $\textit{mini}$ImageNet, which inherit the learngenes evolved different generations.}
    \label{tab:evolution}
\end{table}

In the later stages of evolution, both the parameter quantity of the learngenes and the classification accuracy of neural networks show a convergence trend. The learngenes at this time have significantly surpassed the model trained from scratch and achieved comparable results with the model pre-trained on training classes (i.e., $\mathcal{W}_{\text{train}}$) with only 20\% of the parameters of such pre-trained model. Some networks in the population even exceeded the performance of the pre-trained models.
This fully demonstrates that the knowledge condensed in the learngenes, while limited, is sufficiently common for descendants to adapt to diverse environments.

\subsection{Ablation Experiments}
We set up several degenerated methods: 1) w/o evolution: The learngenes are directly extracted from a pre-trained model based on the structures of our evolved learngenes. 2) w/o tournaments and Gene Pool (tour\&GP): In each generation, winners are randomly selected, and parental learngenes are chosen randomly from the Gene Pool. 3) w/o mutation: the structures of the learngenes keep unchanged throughout the evolutionary process. 4) w/o population: The population size is set to 1. Table \ref{tab:ablation} shows the comparison results. 

The neural circuits extracted from a pre-trained model w/o evolution fails to be the learngenes. Although having the same structures as our learngenes, they skip the process of condensing knowledge. Thus, the extracted knowledge is just a discrete subset of the entire knowledge. In comparison, the evolution process---like biological evolution---continuously selects and inherits the superior learngenes, which gradually condense core knowledge into the learngenes. 

The tournaments and GP maintain the diversity and superiority of learngenes. Without them, the evolution process may get lost and lead to redundant knowledge in the learngenes. Similarly, the population ensures the diversity of learngenes with sufficient candidates for superior learngenes. Indeed, without population, the evolution process degenerates to the continuous learning of different tasks of one network. Mutations enable the learngenes to adjust their structure for better encoding of the core knowledge under selection pressure. 

\begin{table}
    \centering
    \resizebox{\linewidth}{!}{
        \begin{tabular}{@{}lcc|cc@{}}
            \hline
            \multirow{2}{*}{Methods} & \multicolumn{2}{c|}{$\text{\small{VGG11, CIFAR FS}}$} & \multicolumn{2}{c}{\text{\small{ResNet12, $\textit{mini}$Imagenet}}}\\
            & validation & novelty & validation & novelty\\
            \hline
            w/o evolution & 71.19 & 74.30 & 76.12 & 77.10 \\
            w/o tour\&GP & 73.37 & 76.30 & 82.00 & 80.45 \\
            w/o mutation & 74.37 & 75.00 & 81.75 & 80.80 \\
            w/o population & 74.62 & 76.55 & 80.75 & 81.50 \\
            Learngene & \textbf{76.81} & \textbf{78.10} & \textbf{83.75} & \textbf{82.45} \\
            \hline
        \end{tabular}
    }
    \caption{Ablation study results on CIFAR-FS and $\textit{mini}$ImageNet.}
    \label{tab:ablation}
\end{table}

\subsection{Core Knowledge in the Learngenes}
To visually demonstrate the core knowledge condensed by the learngenes, we selected sample images corresponding to novelty classes in $\textit{mini}$ImageNet, and employed CAM~\cite{selvaraju2017grad} to visualize the attention in pre-trained networks (ResNet12 pre-trained on training classes of $\textit{mini}$ImageNet and ResNet50 provided by Pytorch official), as well as those initialized randomly and by the learngenes.

\begin{figure}[tb]
  \centering
  \includegraphics[width=\linewidth]{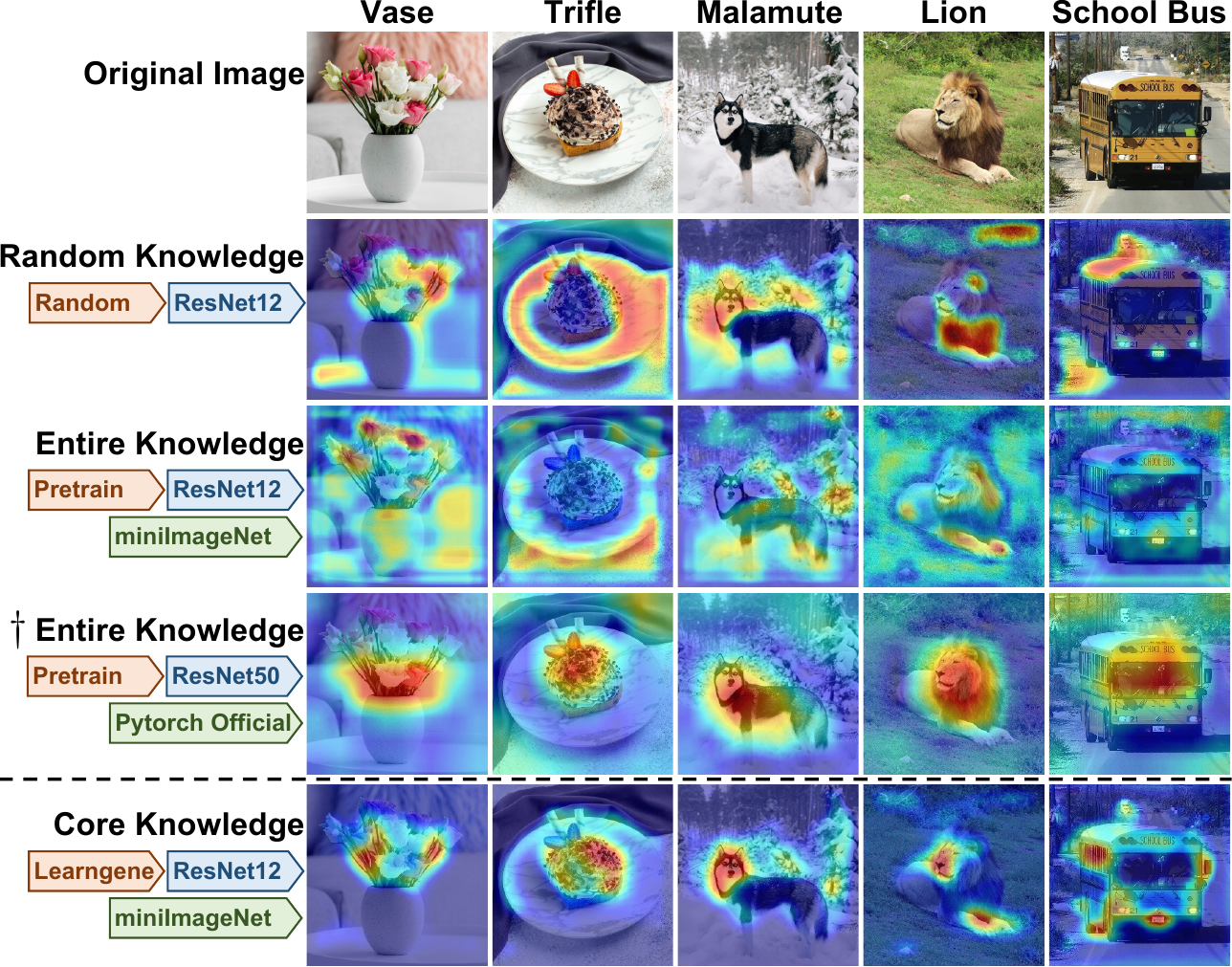}
  \caption{The visualization of core knowledge in the learngenes. All networks have not undergone any learning or fine-tuning.$\quad$ $\dag$ ResNet50 from Pytorch Official was pre-trained on ImageNet, which has included these classes. }
  \label{fig:CAM}
\end{figure}

In Figure \ref{fig:CAM}, the networks with randomly initialized parameters randomly concentrated on certain parts or the whole images. Pre-trained networks transfer the entire knowledge learned before, so they display a broader focus on the whole images. The pre-trained ResNet12 presents divergent attention when facing unknown classes and fails to focus on the object itself that needs to be classified. The ResNet50 pre-trained on the ImageNet has seen these classes before, so it successfully focuses on the objects. But this widespread attention inevitably obtains external information from the background, and may introduce redundancy and affect classification (see Figure \ref{fig:motivation}(a)). In contrast, the learngenes can extract more core knowledge from less training data (only 4\% of the training data used for pre-training ResNet50), which focuses on more local features (i.e., smaller red attention blocks) and therefore has stronger transferability, even when facing unknown classes.

\subsection{Advantages of the Learngenes}

\subsubsection{Instincts}
Instinct rarely appears in the field of AI, as it is a natural ability of organisms brought by genes~\cite{seung2012connectome}. Instincts enable organisms to quickly adapt to the environments with minimal or even no interaction. \cite{feng2023genes} previously disclosed the instincts of RL agents by showing that newborn agents unconsciously move toward rewards. In this study, we manifest the instincts of the networks initialized by the learngenes, which can quickly classify images with minimal gradient descent even with a substantial proportion of randomly initialized neurons. 

Table \ref{tab:instinct} compares the learngene with mainstream network initialization~\cite{he2015delving,zhu2021gradinit,knyazev2021parameter,knyazev2023canwescale} and knowledge transfer~\cite{wang2023semckd,wang2022learngene} methods. 
After 5 training epochs, GradInit, GHN-3, semCKD, and Heru-LG all outperform HeInit, the simplest parameter initialization method.
While these methods achieve higher performance, excessive knowledge transfer can compromise neural network parameter flexibility. In early training stages (i.e., 10, 30, and 50 iterations), GHN-3, Heru-LG, and SemCKD occasionally perform similarly to HeInit and GradInit, as they struggle to adapt to the new classes in a short period. Although these methods bring knowledge to neural networks through direct transfer, distillation, or parameter generation, we do not observe instincts in the neural networks.

In contrast, the learngenes transfer only the core knowledge for the learning of descendant networks. As a result, the networks inheriting the learngenes only require minimal interaction with environments to gain a preliminary understanding of categories and even achieve a degree of classification accuracy. This ability can be understood as the instincts of networks brought by the learngenes. 

\begin{table}
    \centering
    \resizebox{\linewidth}{!}{
        \begin{tabular}{@{}lp{0.75cm}p{0.75cm}p{0.8cm}|p{0.75cm}p{0.75cm}p{0.7cm}@{}}
        \hline
             \multirow{3}{*}{Methods} & \multicolumn{3}{c}{VGG11} & \multicolumn{3}{c}{ResNet12}\\
             & \multicolumn{3}{c}{CIFAR FS, novelty} & \multicolumn{3}{c}{$\textit{mini}$Imagenet, novelty}\\
             \cline{2-7}
             & 30it & 50it & 5ep & 10it & 30it & 5ep\\
             \hline
             HeInit & 16.4$\substack{\text{\tiny{1.6}}}$ & 23.6$\substack{\text{\tiny{1.8}}}$ & 57.2$\substack{\text{\tiny{2.4}}}$ & 8.2$\substack{\text{\tiny{1.3}}}$ & 16.4$\substack{\text{\tiny{1.5}}}$ & 53.4$\substack{\text{\tiny{0.7}}}$\\
             GradInit & 19.7$\substack{\text{\tiny{2.0}}}$ & 23.8$\substack{\text{\tiny{2.0}}}$ & 57.3$\substack{\text{\tiny{0.3}}}$ & 7.5$\substack{\text{\tiny{2.0}}}$ & 17.0$\substack{\text{\tiny{1.1}}}$ & 55.7$\substack{\text{\tiny{1.5}}}$\\
             GHN-2 & 5.0$\substack{\text{\tiny{0.0}}}$ & 5.3$\substack{\text{\tiny{0.5}}}$ & 37.7$\substack{\text{\tiny{0.6}}}$ & 5.5$\substack{\text{\tiny{0.3}}}$ & 5.0$\substack{\text{\tiny{0.2}}}$ & 51.1$\substack{\text{\tiny{0.6}}}$ \\
             GHN-3 & 5.0$\substack{\text{\tiny{0.0}}}$ & 10.7$\substack{\text{\tiny{2.5}}}$ & 56.5$\substack{\text{\tiny{1.6}}}$ & 5.0$\substack{\text{\tiny{0.0}}}$ & 5.1$\substack{\text{\tiny{0.2}}}$ & 68.5$\substack{\text{\tiny{1.4}}}$ \\
             Heru-LG\ddag & 18.1$\substack{\text{\tiny{0.9}}}$ & 25.1$\substack{\text{\tiny{2.2}}}$ & 60.0$\substack{\text{\tiny{0.9}}}$ & 9.2$\substack{\text{\tiny{1.1}}}$ & 17.8$\substack{\text{\tiny{1.7}}}$ & 56.5$\substack{\text{\tiny{1.1}}}$\\
             SemCKD & 17.9$\substack{\text{\tiny{2.7}}}$ & 19.6$\substack{\text{\tiny{4.5}}}$ & 68.8$\substack{\text{\tiny{0.2}}}$ & 13.4$\substack{\text{\tiny{2.7}}}$ & 40.6$\substack{\text{\tiny{1.5}}}$ & 74.8$\substack{\text{\tiny{0.2}}}$\\
             Learngene & \textbf{47.0}$\substack{\textbf{\tiny{1.7}}}$ & \textbf{54.0}$\substack{\textbf{\tiny{1.4}}}$ & \textbf{73.7}$\substack{\text{\tiny{0.2}}}$ & \textbf{21.7}$\substack{\textbf{\tiny{3.0}}}$ & \textbf{48.6}$\substack{\textbf{\tiny{3.6}}}$ & \textbf{79.3}$\substack{\textbf{\tiny{0.7}}}$\\
             \hline
        \end{tabular}
    }
    \caption{Comparison with initialization and knowledge transfer methods on novelty classes of CIFAR-FS and $\textit{mini}$ImageNet. ``it'' represents the parameter update iteration (i.e., 1it represents the optimizer completes one parameter update), and ``ep'' denotes the training epoch. \ddag The size of last block of ResNet12 in Heru-LG is 512.}
    \label{tab:instinct}
\end{table}

\subsubsection{Strong Learning Ability}
Next, we demonstrate that the networks inheriting the learngenes have strong learning abilities even with limited data. We assess the learning ability of the learngenes on the few-shot tasks and compare the learngenes with other few-shot learning algorithms~\cite{finn2017model,sung2018learning,vinyals2016matching,snell2017prototypical,chen2018a,oh2021boil} on CIFAR-FS and $\textit{mini}$ImageNet, averaging the accuracy over 600 tasks.

Table \ref{tab:fewshot} reports the comparison results. Although the networks inheriting the learngenes have a substantial number of randomly initialized parameters, they can still outperform other few-shot learning algorithms that reuse the entire models. Moreover, we leverage the scalability of the learngenes and initialize a network with identical depth but reduced width, denoted by ``-N''. It demonstrates improved accuracy compared to the original network because of less randomly initialized parameters to fit the few-shot tasks. 

\begin{table}
    \centering
    \resizebox{\linewidth}{!}{
        \begin{tabular}{@{}p{1.86cm}p{0.8cm}p{0.8cm}p{0.85cm}|p{0.8cm}p{0.8cm}p{1.0cm}@{}}
            \hline
            \multirow{3}{*}{Methods} & \multicolumn{3}{c}{CIFAR-FS} & \multicolumn{3}{c}{$\textit{mini}$Imagenet}\\
            & \multicolumn{3}{c}{VGG11, 5way} & \multicolumn{3}{c}{ResNet12, 5way}\\
            \cline{2-7}
              & 5shot & 10shot & 20shot & 5shot & 10shot & 20shot \\
            \hline
            MAML & 63.4$\substack{\text{\tiny{0.86}}}$ & 68.2$\substack{\text{\tiny{0.74}}}$ & 70.5$\substack{\text{\tiny{0.77}}}$ & 61.1$\substack{\text{\tiny{0.78}}}$ & 66.4$\substack{\text{\tiny{0.68}}}$ & 68.4$\substack{\text{\tiny{0.62}}}$ \\
            RelationNet & 64.2$\substack{\text{\tiny{0.79}}}$ & 68.9$\substack{\text{\tiny{0.71}}}$ & 72.9$\substack{\text{\tiny{0.71}}}$ & 65.4$\substack{\text{\tiny{0.69}}}$ & 70.3$\substack{\text{\tiny{0.66}}}$ & 72.9$\substack{\text{\tiny{0.63}}}$ \\
            MatchingNet  & 59.9$\substack{\text{\tiny{0.78}}}$ & 63.8$\substack{\text{\tiny{0.78}}}$ & 69.3$\substack{\text{\tiny{0.81}}}$ & 66.3$\substack{\text{\tiny{0.66}}}$ & 70.9$\substack{\text{\tiny{0.63}}}$ & 74.7$\substack{\text{\tiny{0.59}}}$\\
            ProtoNet & 65.9$\substack{\text{\tiny{0.85}}}$ & 69.3$\substack{\text{\tiny{0.79}}}$ & 73.1$\substack{\text{\tiny{0.69}}}$ & 66.5$\substack{\text{\tiny{0.71}}}$ & 72.4$\substack{\text{\tiny{0.60}}}$ & 74.9$\substack{\text{\tiny{0.59}}}$ \\
            Baseline++  & 64.9$\substack{\text{\tiny{0.78}}}$& 71.3$\substack{\text{\tiny{0.73}}}$  & 75.3$\substack{\text{\tiny{0.67}}}$& 67.5$\substack{\text{\tiny{0.67}}}$ & 74.0$\substack{\text{\tiny{0.60}}}$ & 78.2$\substack{\text{\tiny{0.51}}}$ \\
            BOIL & 68.3$\substack{\text{\tiny{0.76}}}$& 71.5$\substack{\text{\tiny{0.71}}}$& 72.9$\substack{\text{\tiny{0.64}}}$ & 67.8$\substack{\text{\tiny{0.69}}}$ & 72.4$\substack{\text{\tiny{0.63}}}$ & 75.0$\substack{\text{\tiny{0.60}}}$ \\
            \hline
            Learngene & 69.9$\substack{\text{\tiny{0.78}}}$ & 75.5$\substack{\text{\tiny{0.69}}}$ & 78.5$\substack{\text{\tiny{0.63}}}$ & 69.4$\substack{\text{\tiny{0.71}}}$ & 75.4$\substack{\text{\tiny{0.61}}}$ & 80.2$\substack{\text{\tiny{0.52}}}$\\
            Learngene-N & \textbf{70.5$\substack{\text{\tiny{0.73}}}$} & \textbf{76.6$\substack{\text{\tiny{0.65}}}$} & \textbf{80.5$\substack{\text{\tiny{0.58}}}$} & \textbf{71.3$\substack{\text{\tiny{0.70}}}$} & \textbf{76.8$\substack{\text{\tiny{0.59}}}$} & \textbf{81.7$\substack{\text{\tiny{0.53}}}$} \\
            \hline
        \end{tabular}
    }
    \caption{Accuracy of few-shot classification. ``-N'' indicates narrower networks than normal ones. }
    \label{tab:fewshot}
\end{table}

\subsubsection{Scalability and Adaptability}
Besides initializing narrower networks for few-shot tasks, the learngenes exhibit broader scalability of initializing networks with varying depths, widths, and architectures. The core knowledge in the learngenes also demonstrates strong adaptability when applied to fine-grained datasets. Table \ref{tab:scalability} is the results of different networks on four fine-grained datasets. 

Tables \ref{tab:scalability} shows that the learngenes (i.e., $\mathcal{G}_{\text{vgg}}$ and $\mathcal{G}_{\text{resnet}}$) can initialize the networks with different depths (e.g., 11, 16, and 19 for VGG; 12 and 18 for ResNet), which all outperforms those learning from scratch. The learngenes demonstrate remarkable scalability to depth and successfully transfer the core knowledge to fine-grained classification tasks. 

\begin{table}
    \centering
    \resizebox{\linewidth}{!}{
        \begin{tabular}{@{}p{0.05cm}p{0.6cm}p{0.42cm}p{0.42cm}p{0.42cm}p{0.42cm}p{0.48cm}|p{0.42cm}p{0.42cm}p{0.42cm}p{0.48cm}}
            \hline
            \multicolumn{2}{c}{\multirow{2}*{Arch}} & \multicolumn{5}{c}{VGG} & \multicolumn{4}{c}{ResNet}\\
            \cline{3-11}
             & & 11 & $\text{11}_\text{--N}$ & $\text{11}_\text{--W}$ & $\text{16}$ & $\text{19}$ & $\text{12}$ & $\text{12}_\text{--N}$ & $\text{12}_\text{--W}$ & \text{18} \\    
            \hline
            \multirow{3}{*}{\rotatebox{90}{Flower}} & $\mathcal{S}$ & 51.4 & 49.2 & 53.2 & 47.3 & 41.8 & 56.6 & 56.2 & 56.6 & 56.2 \\
            & $\mathcal{G}_{\text{vgg}}$ & \underline{69.3} & \underline{66.9} & \underline{67.0} & \underline{72.0} & \underline{74.5} & 69.0 & 66.2 & 66.2 & 63.2 \\
            & $\mathcal{G}_{\text{resnet}}$ & 59.2 & 56.9 & 57.7 & 69.8 & 60.7 & \underline{77.3} & \underline{\textbf{79.8}} & \underline{74.6} & \underline{77.0} \\
            \hline
            \multirow{3}{*}{\rotatebox{90}{CUB}} & $\mathcal{S}$ & 70.8 & 66.9 & 73.3 & 66.6 & 69.3 & 64.0 & 63.7 & 64.0 & 67.6 \\
            & $\mathcal{G}_{\text{vgg}}$ & \underline{79.7} & \underline{78.7} & \underline{81.6} & 82.7 & \underline{\textbf{84.2}} & 76.7 & 75.2 & 76.7 & 78.2  \\
            & $\mathcal{G}_{\text{resnet}}$ & 77.9 & 76.0 & 79.6 & \underline{83.5} & 81.2 & \underline{81.6} & \underline{82.2} & \underline{80.3} & \underline{83.6} \\
            \hline
            \multirow{3}{*}{\rotatebox{90}{Cars}} & $\mathcal{S}$ & 86.0 & 81.1 & 88.2 & 87.1 & 89.2 & 80.1 & 74.4 & 80.4 & 82.5\\
            & $\mathcal{G}_{\text{vgg}}$ & \underline{92.3} & \underline{91.4} & \underline{93.0} & 94.6 & \underline{95.3} & 92.8 & 91.7 & 93.0 & 93.6\\
            & $\mathcal{G}_{\text{resnet}}$ & 89.8 & 88.1 & 91.3 & \underline{95.2} & 94.7 & \underline{95.6} & \underline{95.9} & \underline{95.4} & \underline{\textbf{96.1}}\\
            \hline
            \multirow{3}{*}{\rotatebox{90}{Food}} & $\mathcal{S}$ & 80.7 & 79.7 & 80.3 & 79.5 & 76.2 & 84.6 & 84.2 & 85.3 & 87.6\\
            & $\mathcal{G}_{\text{vgg}}$ & \underline{85.5} & \underline{84.7} & \underline{84.8} & \underline{87.3} & \underline{86.9} & 88.2 & 87.3 & 88.3 & 89.0 \\
            & $\mathcal{G}_{\text{resnet}}$ & 84.7 & 83.1 & 84.4 & 86.5 & 84.3 & \underline{89.5} & \underline{89.1} & \underline{89.7} & \underline{\textbf{90.6}}\\
            \hline
        \end{tabular}
    }
    \caption{Accuracy of fine-grained classification on Oxford 102 Flower, CUB-200-2011, Standford Cars, and Food-101 with diverse structured networks. ``$\mathcal{S}$'' denotes training from scratch. ``$\mathcal{G}_{\text{vgg}}$'' and ``$\mathcal{G}_{\text{resnet}}$'' denote inheriting the learngenes extracted from VGG11 and ResNet12 in Section \ref{sec:extract learngene}, respectively. ``-N/W'' indicates narrower/wider depth. More details can be found in \textbf{Appendix C}.}
    \label{tab:scalability}
\end{table}

The learngenes also effectively initialize the networks with narrower (N) or wider (W) widths than standard ones to improve the performance. For example, $\mathcal{G}_{\text{vgg}}$ initializes a wider VGG11-W that outperforms a standard VGG11 on CUB (81.6\% vs. 79.7\%), and $\mathcal{G}_{\text{resnet}}$ initializes a narrower ResNet12-N with better performance than a standard ResNet12 on Flower (79.8\% vs. 77.3\%).

The core knowledge can also be transferred across architectures. For example, $\mathcal{G}_{\text{vgg}}$ can initialize ResNet18 with better performance than learning from scratch.
Despite potential knowledge incompatibility caused by architectural disparities, the flexibility of the learngenes still surpasses that of pre-trained models, which impose stricter architecture requirements and have less scalability.

\section{Conclusion}
In this study, motivated by the efficient knowledge transfer in nature via genes, we introduce a novel approach for knowledge transfer in neural networks, that is, condensing the core knowledge into the learngenes and transferring it by inheriting the learngenes. 
We refine the structures of the learngenes as neural circuits, and propose the Genetic Transfer Learning (GTL), a framework for the evolution of neural networks and transfer of the learngenes in supervised learning. 
The learngenes extracted in our experiments are discrete neural circuits with continuous mappings, which can transfer core knowledge in an efficient way. Additionally, the learngenes bring several characteristics to neural networks, such as instincts and strong learning ability, and display scalability and adaptability to diverse network structures and training data in downstream tasks. Overall, we copy the biological knowledge transfer into neural network and provide an alternative way for knowledge transfer via the learngenes, emphasizing the transfer of core knowledge.
\section*{Acknowledgments}
We sincerely thank Wenqian Li, Congzhi Zhang and Jiawei Peng for the helpful discussion, and thank macrovector, brgfx, pikisuperstar, KamranAydinov, wirestock, freepic.diller and freepik for designing some figures.
This research is supported by the National Key Research \& Development Plan of China (No. 2018AAA0100104), the National Science Foundation of China (62125602, 62076063) and Xplorer Prize.

\bibliographystyle{named}
\bibliography{ijcai24}
\onecolumn
\newpage
\twocolumn
\appendix
\section{Form of the Learngenes in ResNets}
\begin{figure*}[tb]
    \centering
    \includegraphics[width=\linewidth]{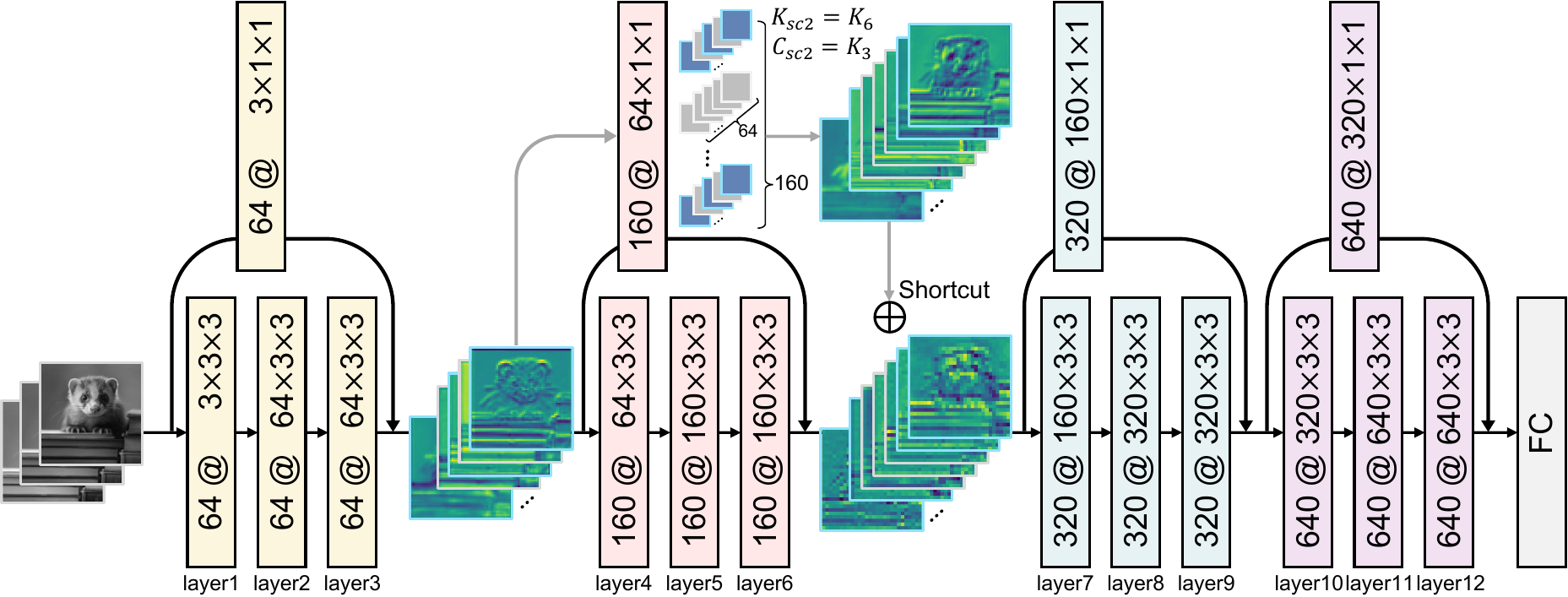}
    \caption{The form of the learngenes in ResNet12, where the kernels in skip connection layers are also integrated as components of the learngenes.}
    \label{fig:gene_resnet}
\end{figure*}

We have demonstrated the form of the learngenes in CNNs (Section 3.1), which are several complete neural circuits in the unit of channels within kernels. In contrast to conventional CNNs (e.g., VGG), ResNets incorporate skip connection layers into their architectures, typically comprising 1$\times$1 convolution kernels that facilitate the transfer of feature maps across multiple layers. To ensure the learngenes retain their structures as complete neural circuits with continuous mappings in ResNets, we additionally extracted channels from skip connection layers to serve as components of the learngenes, as depicted in Figure \ref{fig:gene_resnet}. For a skip connection layer $l_{\text{sc}}$ positioned between $l_i$-th and $l_j$-th layer ($l_j>l_i$), the number of kernels and their corresponding channels within kernels is $n_K^{l_{\text{sc}}}=n_K^{l_j}$ and $n_C^{l_{\text{sc}}}=n_K^{l_i}$, respectively. Consequently, the indices of the kernels and channels corresponding to the learngenes in $l_{\text{sc}}$ are $K_{l_{\text{sc}}}=K_{l_j}$ and $C_{l_{\text{sc}}}=K_{l_i}$.

\section{Details of Inheriting the Learngenes}
The learngenes exhibit scalability, enabling the initialization of descendant networks with varying width, depth, and architectures, as demonstrated in Section 3.3 and Section 4.5 for methods and results, respectively.

\begin{figure*}[tb]
    \centering
    \includegraphics[width=\linewidth]{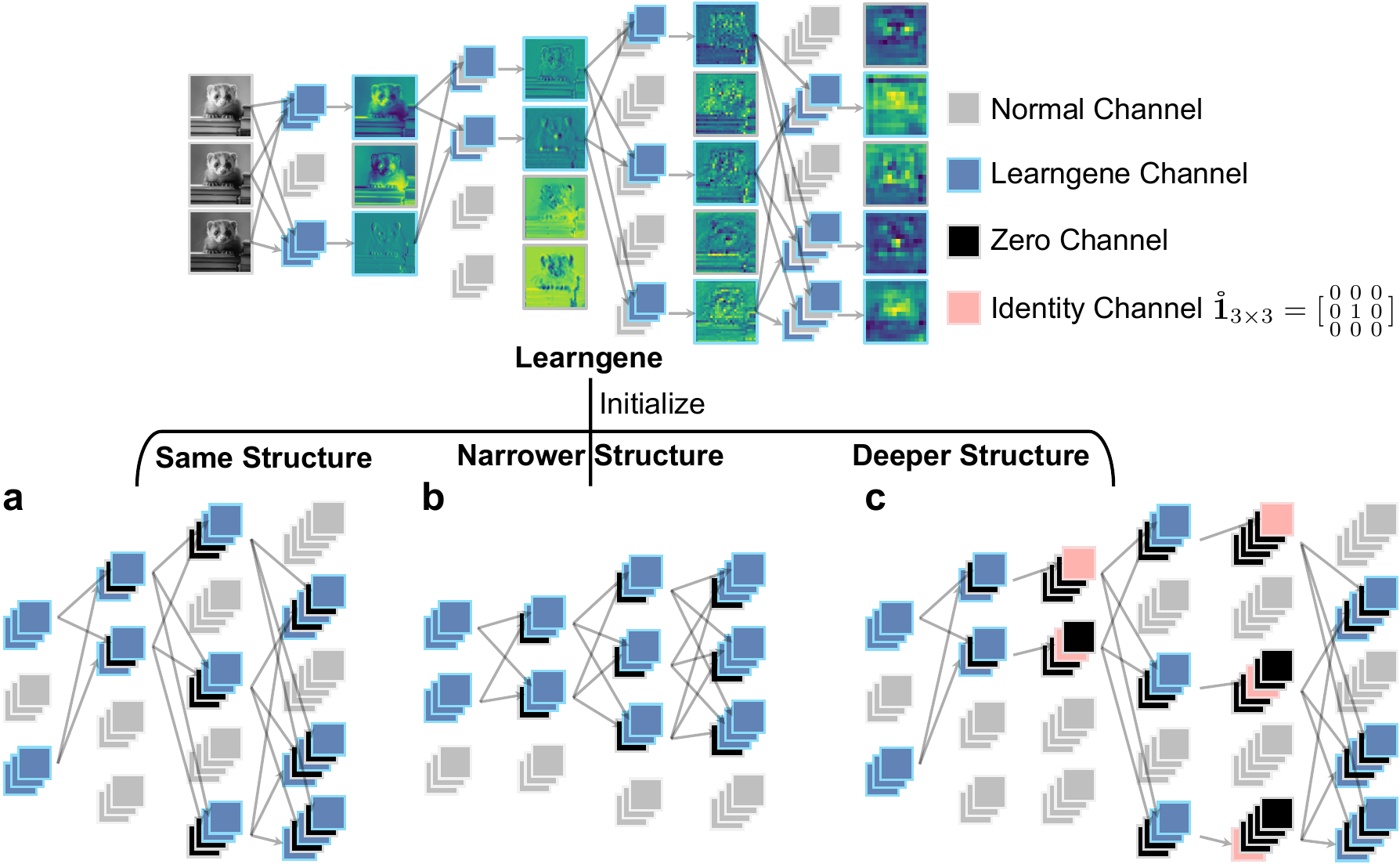}
    \caption{The learngenes exhibit scalability, enabling the initialization of networks with (a) the same structures and the flexibility to initialize networks with (b) narrower/wider and (c) deeper structures. The normal channels in networks are randomly initialized, while the zero channels and identity channels are initialed by \textbf{0} (i.e., zero matrix) and $\mathbf{\mathring{1}}$, respectively.}
    \label{fig:scalability}
\end{figure*}

As depicted in Figure \ref{fig:scalability}, when the descendant network shares the same structure as the ancestry network (Figure \ref{fig:scalability}a), a straightforward replacement of randomly initialized kernels is performed based on the indices of kernels and channels corresponding to the learngenes. Additionally, missing channels within kernels are filled with $\mathbf{0}$ (depicted as black kernels in Figure \ref{fig:scalability}).

For descendant networks with widths narrower/wider than the ancestry networks, the indices of kernels and channels in the $l$-th layer of the learngene should be sorted to $K_l'=[1, |K_l|]$ and $C_l'=[1, |C_l|]$, respectively, while ensuring $K_l' = C_{l+1}'$ (where $|\cdot|$ denotes the size of the set) to prevent indices from going out of range (Figure \ref{fig:scalability}b).

In cases where descendant networks have greater depth, \textit{partial identity mapping} layers $l_{\text{pim}}$ are added to the learngenes (Figure \ref{fig:scalability}c). These layers not only align the number of layers in the learngenes, but also ensure the continuous mapping of core features within the learngenes.

\section{Experimental Details}
\begin{figure*}[tb]
    \centering
    \includegraphics[width=\linewidth]{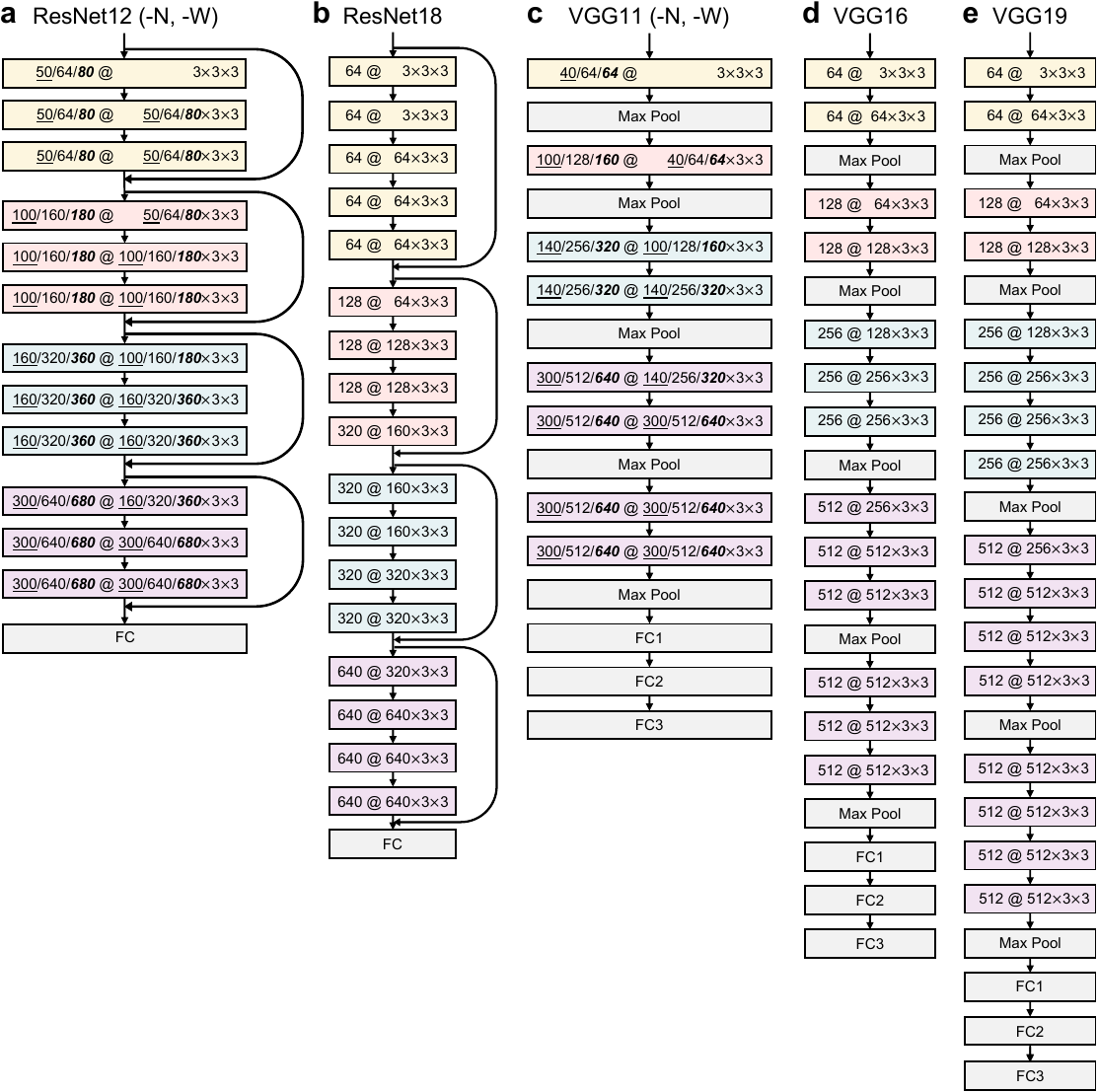}
    \caption{The structure of networks employed in our experiments. In \textbf{a} and \textbf{c}, the underscored numbers (e.g., \underline{50}) indicate the kernel/channel numbers in narrower networks (i.e., VGG11-N and ResNet12-N), while the bold and italicized numbers (e.g., \textbf{\textit{80}}) represent the kernel/channel numbers in wider networks (i.e., VGG11-W and ResNet12-W). The skip connection layers in ResNet18 are slightly modified in quantity and placement to enhance the inheritance of the learngenes.}
    \label{fig:net_structure}
\end{figure*}

\begin{figure*}[tb]
    \centering
    \includegraphics[width=\linewidth]{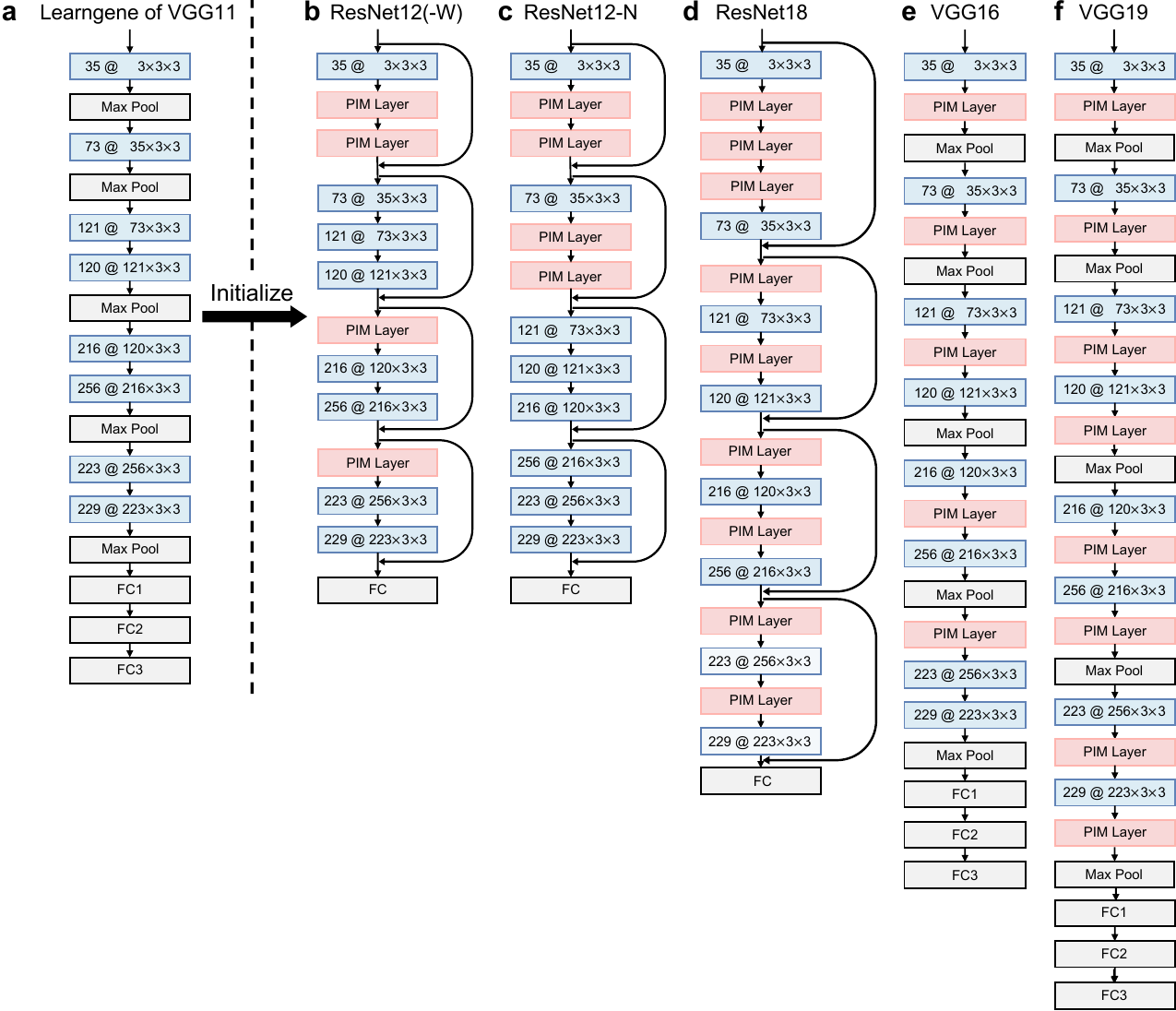}
    \caption{The initialization details of the learngenes (extracted from VGG11) for ResNet12(-N, -W), ResNet18, VGG16, and VGG19. Since the learngenes extracted from VGG11 lack skip connection layers, when initializing ResNets, the relevant kernels and channels in the skip connection layer $l_{\text{sc}}$ (i.e., kernels and channels in $K_{\text{sc}}$ and $C_{\text{sc}}$) are initialized with \textbf{0}.}
    \label{fig:inherit_vgg}
\end{figure*}

\begin{figure*}[tb]
    \centering
    \includegraphics[width=\linewidth]{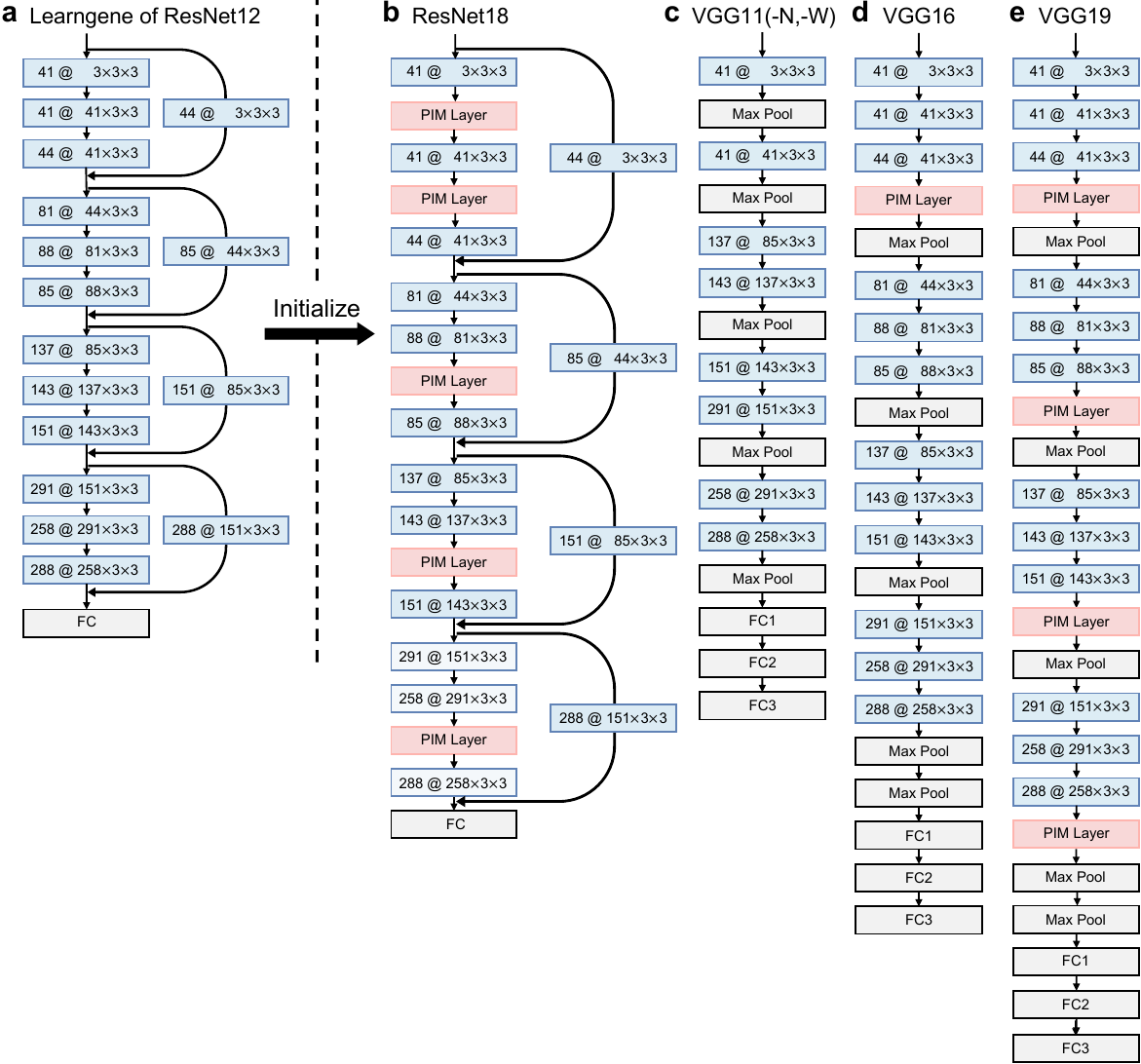}
    \caption{The initialization details of the learngenes (extracted from ResNet12) for ResNet18, VGG11(-N, -W), VGG16, and VGG19. The positions of the max pooling layers in VGG16 and VGG19 are slightly modified when inheriting the learngenes extracted from ResNet12.}
    \label{fig:inherit_res}
\end{figure*}

In our evolutionary experiments with neural networks (Section 4.2), we employed VGG11 (Figure \ref{fig:net_structure}c) as the structure of the population for evolving on CIFAR-FS and ResNet12 (Figure \ref{fig:net_structure}a) for \textit{mini}ImageNet. Following 250 generations of evolution, we successfully extracted the learngenes from the VGG11 (Figure \ref{fig:inherit_vgg}a) and ResNet12 (Figure \ref{fig:inherit_res}a). Remarkably, the parameters of these learngenes comprise only approximately 20\% of the total network parameters.

Subsequently, we conducted experiments on fine-grained datasets to demonstrate the scalability of the learngenes (Section 4.5). Utilizing the extracted learngenes, we initialized networks with varying width (e.g., VGG11-N, VGG11-W, ResNet12-N, and ResNet12-W (Figure \ref{fig:net_structure}a,c)), depth (e.g., VGG16, VGG19, and ResNet18 (Figure \ref{fig:net_structure}b,d,e)), and even different architectures (e.g., VGG$\rightleftharpoons$ResNet). Figure \ref{fig:inherit_vgg} and Figure \ref{fig:inherit_res} show the positions of the introduced \textit{partial identity mapping} layers, denoted as PIM Layer, within the learngenes.

\end{document}